\documentclass[conference]{IEEEtran}
\IEEEoverridecommandlockouts
\usepackage{cite}
\usepackage{amsmath,amssymb,amsfonts}
\usepackage{algorithmic}
\usepackage{graphicx}
\usepackage{textcomp}
\usepackage{xcolor}
\usepackage{placeins}
\usepackage{amsmath}
\usepackage{caption}
\usepackage{subcaption}
\usepackage{tabularx,booktabs, longtable}

\newcolumntype{B}{>{\centering\arraybackslash}X} 
\def\BibTeX{{\rm B\kern-.05em{\sc i\kern-.025em b}\kern-.08em
T\kern-.1667em\lower.7ex\hbox{E}\kern-.125emX}}
\makeatletter
\newcommand*\titleheader[1]{\gdef\@titleheader{#1}}
\AtBeginDocument{%
\let\st@red@title\@title\def\@title{%
\bgroup\normalfont\large\centering\@titleheader\par\egroup
\vskip1.5em\st@red@title}
}
\makeatother
\makeatletter
\newcommand{\linebreakand}{%
\end{@IEEEauthorhalign}
\hfill\mbox{}\par
\mbox{}\hfill\begin{@IEEEauthorhalign}
}
\author{\IEEEauthorblockN{1\textsuperscript{st} Mohana Sri S}
      \IEEEauthorblockA{\textit{Department of Electronics and}\\
\textit{Communication Engineering}\\
      \textit{Sri Sai Ram Engineering College}\\
      Chennai, 
      India\\
      s.mohanasri7@gmail.com\\
      }
\and
\IEEEauthorblockN{2\textsuperscript{nd} Swethaa S}
      \IEEEauthorblockA{\textit{Department of Electronics and}\\
\textit{Communication Engineering}\\
      \textit{Sri Sai Ram Engineering College}\\
      Chennai, 
      India\\
      swethaasankarraman@gmail.com\\
      }
\and
\IEEEauthorblockN{3\textsuperscript{rd} Aouthithiye Barathwaj SR Y}
      \IEEEauthorblockA{\textit{Department of Electrical and}\\
\textit{Electronics Engineering}\\
      \textit{Sri Sai Ram Engineering College}\\
      Chennai, 
      India\\
      sec19ee038@sairamtap.edu.in\\
      }
\linebreakand
\IEEEauthorblockN{4\textsuperscript{th} Sai Ganesh CS}
      \IEEEauthorblockA{\textit{Technology and Operations}\\
      \textit{Genik Technologies Private Limtied}\\
      Chennai, 
      India\\
      saiganeshcs@geniktech.com\\
      }
}
\makeatother
\makeatletter\let\old@ps@headings\ps@headings
\let\old@ps@IEEEtitlepagestyle\ps@IEEEtitlepagestyle
\def\confheader#1{%
\def\ps@headings{%
\old@ps@headings%
\def\@oddhead{\strut\hfill#1\hfill\strut}%
\def\@evenhead{\strut\hfill#1\hfill\strut}%
}%
\def\ps@IEEEtitlepagestyle{%
\old@ps@IEEEtitlepagestyle%
\def\@oddhead{\strut\hfill#1\hfill\strut}%
\def\@evenhead{\strut\hfill#1\hfill\strut}%
}%
\ps@headings%
}
\makeatother

\usepackage[pscoord]{eso-pic}
\newcommand{\placetextbox}[3]{
\setbox0=\hbox{#3}
\AddToShipoutPictureFG*{ \put(\LenToUnit{#1\paperwidth},\LenToUnit{#2\paperheight}){\vtop{{\null}\makebox[0pt][c]{#3}}}
}
}
\placetextbox{.23}{0.055}{\small{}}
\begin{document}
\title{Intelligent Debris Mass Estimation Model for Autonomous Underwater Vehicle}
\maketitle
\begin{abstract}Marine debris have detrimental effects on marine life including entanglement and ingestion by marine organisms. Estimating the mass of marine debris is essential to understand the severity and depths of its impact on marine aquaculture. The methodology involved conducting a comparative analysis of all YOLO algorithms and their performance that enables future researchers to study and select the appropriate model for their specific needs. In this paper, we use instance segmentation to calculate the area of individual objects within an image, using YOLOv7 in Roboflow. YOLOv7 is a fast and accurate object detection algorithm, capable of processing up to 160 frames per second with the highest accuracy of 56.8 \% among well-known object detectors. To perform instance segmentation, we use YOLOv7 in Roboflow to generate a set of bounding boxes for each object in the image with a class label and a confidence score for every detection. A segmentation mask is then created for each object by applying a binary mask to the object's bounding box. The masks are generated by applying a binary threshold to the output of a convolutional neural network trained to segment objects from the background. Finally, refining the segmentation mask for each object is done by applying post-processing techniques such as morphological operations and contour detection, to improve the accuracy and quality of the mask. The process of estimating the area of instance segmentation involves calculating the area of each segmented instance separately and then summing up the areas of all instances to obtain the total area. The calculation is carried out using standard formulas based on the shape of the object, such as rectangles and circles. In cases where the object is complex, the Monte Carlo method is used to estimate the area. This method provides a higher degree of accuracy than traditional methods, especially when using a large number of samples.\end{abstract}
\begin{IEEEkeywords}
Computer vision, Debris , Marine debris, Debris Mass Estimation, Autonomous Underwater Vehicles, Yolo algorithms ,Instance segmentation, Machine learning.
\end{IEEEkeywords}
\section{INTRODUCTION}
Coastal pollution causing by marine debris owe a disastrous outgrowth on ecosystems, human and marine lives. Marine debris has a profound and serious impact on marine ecosystems. Marine lives are harmed through ingestion, entanglement, and habitat destruction by marine debris mistaking it for food leading to their internal injuries and even death. Debris smother coral reefs and other habitats that can disrupts ecosystems and  food chains leading to cascading ecological effects. Marine Debris abide for decades or even centuries in the environment resulting in a tedious presence of pollutants in marine aquaculture. The consequences of marine debris elongate beyond immediate carnage. The Chronic hazard to debris leads to physiological and behavioral changes in marine organisms certainly debilitating their reproductive systems and overall wellness. The outcomes of marine debris extend beyond wildlife embracing the hidden adverse impacts on human health.
The United Nation's 2023 Sustainable Development Goals (SDG) submission highlights annual plastic production has been surging from 1.5 million tonnes in the 1950s to a surprising 288 million tonnes in 2012 over the past six decades with the East Asia, Europe, and North America as the major contributers. Global approximations suggest that in 2010 about 275 million tonnes of waste were generated by 192 coastal countries with 4.8 to 12.7 million tonnes ending up in marine environments [1]. According to the United Nations Environment Programme (UNEP), 60 major cities of India produces 15,343 tonnes of waste and it is disposed into the South Asian seas daily. The Data from the 2022 Swachh Sagar, Surakshit Sagar campaign disclosed Indian coastline accumulates around 0.98 metric tonnes of debris per kilometer. The United Nations Environment Programme (UNEP)'s fourth meeting in November 2020 exposed 90 million plastic medical masks contributed further to the marine debris crisis as the effect of COVID-19 pandemic. As per a report by THE HINDU news updated on November 18, 2018, Bindu Sulochanan, a marine ecologist at Mangalore’s Central Marine Fisheries Research Institute (CMFRI), discovered plastic in the stomachs of various species since 2009. The Central Marine Fisheries Research Institute (CMFRI)’s researchers found plastic in species like mackerel near Mangalore, yellowfish tuna near Kochi, and anchovies off Alappuzha's coast. In 2014, Gujarat's Sasan Gir Forest Department examined a deceased Longman’s Beaked Whale weighing a ton, discovering four significant plastic bags blocking its digestive system, bringing out  the severe outcomes of plastic debris on marine life [2]. On October 31, 2020, on the Murcian coast in Spain a infected sperm whale having consumed 64 pounds of plastic waste, including debris like ropes and net was discovered. This event recalls a 2016 occurrence where fishing gear and a car engine cover were found within the stomachs of beached sperm whales along Germany's North Sea coast[3]. The article "Ghosts of the Gulf: Marine Debris a Threat to Corals in the Gulf of Mannar," by Aathira Perinchery on 18 January 2021 in The India Mongabay news discussed abondoned fishing gear and plastic debris endanger Gulf of mannar’s corals specifically delicate Acropora causing disintegrate and harm. Also resilient Thoothukudi coral reefs have recovered from bleaching events. The Lakshadweep reefs in the Arabian Sea also shows similar concerns as highlighted by Scientist Rohan Arthur [4]. The scientific study titled "Microplastic Pollution in Seawater and Marine Organisms across the Tropical Eastern Pacific and Galápagos" by Alonzo Alfaro-Núñez, published on February 25, 2022, revealed marine debris occupied 4,53,000 square kilometers in the Tropical Eastern Pacific and Galápagos. 240 specimens of 16 diverse species of fish, squid, and shrimp, which are consumed by humans contained microplastic particles are found on the coast. Among the species studied, carnivorous organisms displayed the highest microplastic presence at 77\% in their digestive systems, followed by planktivores at 63\% and detritivores at 20\%. The giant squid, Dosidicus gigas, exhibited the highest prevalence at 93\%, followed by Alopias pelagicus and Coryphaena hippurus, both at 87\%. These findings highlighted the alarming levels of microplastic pollution along the Pacific Equator Coast and marks the first documented case of microplastic presence in marine organisms that are consumed by humans in that region [5].The Great Pacific Garbage Patch is located in the North Pacific between California and Hawaii, is a vast region of marine debris. Researches have detected up to 7,50,000 plastic pieces per square kilometer (or 1.9 million per square mile) and more than 200,000 pieces of debris per square kilometer (520,000 per square mile) in the regions of Atlantic garbage patch [6]. A British Broadcasting Corporation (BBC) article from December 5, 2021, reported marine creatures are founded alive in 90\% of the debris along this coast. Dr Linsley Haram of the Smithsonian, Environmental Research centre, study emphasizes that the lasting habitat are formed by enduring plastics of 79000 tonnes [7]. This alarming pattern is predicted to almost triple by the year 2040 and could surge dramatically to 33 billion tons by 2050.

Hence to eradicate the prevailing alarming situations marine debris monitoring plays a vital role in evaluating pollution characteristics for determining suitable action to pollution control. Artificial intelligence technologies like computer vision in marine science have great potential for determining marine debris. Using computer vision for the detection and identification of plastic objects in the marine environment helps to uncover accurate extent that is essential for proposing corrective actions. This paper introduces the integration of Inhibiting (Artificial intelligence) AI algorithm alongside AUVs equipped with advanced sensors, cameras, and imaging systems for the purpose of estimating marine debris mass through instance segmentation techniques, employing different versions of  YOLO (You Only Look Once) algorithms. Further the torque required by the AUVs motor to pull the debris is also determined. The collected data are uploaded to the cloud platform and the output is displayed in the form of website. It helps in accurately detecting the amount of debris present in aquatic environments enabling exact measurements of debris size, shape and density aiding more precise mass calculations allowing for continuous and real-time monitoring of debris accumulation evaluating the efficacy of cleanup ensuring cleaner future for our oceans. This paper explains the techniques employed for mass estimation and embedding the AI algorithm in the AUV along with  a thorough exploration of the conducted experiments and their corresponding findings.

\section{LITERATURE REVIEW}
Autonomous underwater vehicles (AUVs) effectively contribute to detect and remove marine debris. Numerous deep-learning algorithms are evaluated to detect visually the marine debris. A large dataset of debris is annotated and trained with convention neural networks for object detection. To fit the algorithm for real-time application the model is evaluated on various platforms [8]. Achieving rapid detection , identification of autonomous underwater vehicles (AUVs) and cooperative objects remains a challenge due to the complexities of the underwater environment. This paper employs the YOLO (You Only Look Once)  algorithm to identify underwater debris as it have good impact on target detection accuracy and recognition speed. Image enhancement techniques like histogram equalization and the Contrast-limited adaptive histogram equalization (CLAHE) algorithm are applied to enhance the images. These enhanced images are trained on both the yolov2 and yolov3 networks. The experiments outcome revealed the combination of YOLOv3 network and the Contrast-limited adaptive histogram equalization (CLAHE) algorithm effectively fulfills the criteria for rapid and accurate recognition in the detection and identification of underwater vehicles [9]. To overcome the difficulties in assessing the characteristics of marine debris on hard to reach places a method was developed using a segmentation model and images obtained by unmanned aerial vehicle (UAV)s. The conventional statistical estimation method overestimated coastal debris items at 6741 (±1960.0) more than the mapping method. The developed method offered segmentation model with a F1-score of approximately 0.74 for estimating a covered area around 177.4 m² [10]. In underwater environments, the conventional stereo visual SLAM (simultaneous localization and mapping) technique relies on trackable features for camera positioning and mapping but these dependable points are often absent. To enhance the precision of vision-based localization systems in underwater environments an innovative approach is introduced. By Integrating point and line data it investigated a stereo point and line SLAM (PL-SLAM) algorithm that enhances localization and validated through precise experiments using an AR-marker [11]. To estimate PMD (Plastic Marine Debris) volumes a new strategy that combines unmanned aerial vehicle (UAV) surveys with deep learning-based image processing was introduced. A 3D model and orthoscopic beach image are formed using Structure from Motion software by employing data from UAVs. It enabled image-based edge detection for Physical Media-Dependent(PMD) volume calculation. It offered a rapid, precise, and objective alternative to subjective beach surveys by providing (Physical Media-Dependent ) PMD volume estimation accuracy below 5\% [12]. This paper focused on optimizing AUV vision for real-time and low-light object detection. It achieved efficiency enhancements in state-of-the-art object detectors EfficientDets by up to 2.6\% AP across different levels without improving the Graphics processing unit GPU latency. It also introduced a new dataset for detecting in-water debris and trained the improved detectors. The effectiveness and speed of two low-light underwater image enhancement strategies are evaluated [13]. The YOLO algorithm is enhanced by post-processing techniques such as non-maximum suppression (NMS) to refine the detected object bounding boxes, removing duplicates or overlapping detections and optimizing the network architecture. This proposed method demonstrated encouraging outcomes in debris detection accuracy and computational efficiency. One limitation of the approach is that it depends on clear and high-resolution underwater images for ideal performance which is not always be feasible in real-world scenarios and should focus on addressing the challenges posed by low visibility and varying environmental conditions in underwater debris detection [14]. In this article the authors proposed a multi-stage algorithm that combined image preprocessing, feature extraction, and classification techniques. One limitation of the approach is that it struggled with detecting small or partially buried debris objects. Future research should explore techniques for improving the algorithm's performance in these challenging scenarios to enhance underwater debris detection capabilities [15].
\section{METHODOLOGIES}
The AUV is launched by checking the sensors, cameras, propulsion and communication systems working status. To accelerate the autonomous movement under the ocean, Navigation and control systems are initialized. After being launched in a suitable platform, it records images and videos of the submerged environment and classify them. [16] AUV’S like the REMUS 6000 a subsidiary of Kongsberg Maritime, designed by the Naval Oceanographic Office, uses Dual-frequency Side-Scan Sonar and Synthetic Aperture Sonar (SAS) to assemble a 2D image of the seafloor via sound waves and the objects resting on it. It uses standard sensors and camera with high intensity for capturing high resolution images of the marine debris. [17] Iver4 900 Unmanned underwater vehicles (UUV) developed and manufactured by L3Harris OceanServer use Dual Frequency Sonar Side Scan, to capture and process detailed images of marine debris by detecting the echoes emitted from the signals and bouncing off from the seafloor. The Inertial Navigation System (INS) offers accurate positioning, orientation, and velocity data, ensuring precise AUV navigation for targeted marine debris detection. The Sound Velocity Profiler (SVP) sensor measures water’s speed of sound facilitating precise marine debris identification through correct signal interpretation. Similarly [18] The Bluefin-21 AUV utilizes its standard payloads, side scan sonar ,sub-bottom profiler, and the multi-beam echo-sounder to detect marine debris. These sensors captures the detailed images of the seafloor and underwater objects potentially identifying marine debris based on its distinct acoustic signature and shape. The camera system enhancing marine debris detection by capturing high-resolution black and white images of the seafloor and any debris present. These AUVs detect the type of marine debris by implementing the computer vision techniques to the data stored on the AUV’s onboard data storage system that can be visualized and analyzed in real-time or post-mission using specialized software tools. [19] The EvoLogics’s SONOBOT 5 uncrewed surface vehicle which came out in March 2023, employs single-beam/multibeam echosounders, side-scan sonar, and High Definition (HD) camera to capture images and videos. Notably, it introduces Object Recognition (OR), an onboard AI-based system that swiftly identifies and highlights objects from raw side-scan sonar or video output, operating even during mission. Neural network algorithms handle sonar data processing in real-time on dedicated hardware. A cloud-based ecosystem enhances Object Recognition (OR), providing updates and allowing user dataset uploads to train for new object recognition. Raw data is analyzed onboard instantly, showcasing an integration of advanced technology that advances marine exploration, surveillance, and object recognition.

In this paper to get more accurate classification of debris and to estimate their mass, instance segmentation is employed to calculate the area of individual objects within an image, leveraging different iterations of YOLO algorithms within the Roboflow framework. In instance segmentation, the area of each mask is determined by counting the number of pixels in the mask. This involves creating a binary mask for each object in the image, where the pixels that correspond to the object are set to 1, and the background pixels are set to 0. After creating the binary mask, the area of the object is calculated by counting the number of pixels in the mask that have a value of 1. This pixel count corresponds to the area of the object in pixels. The size of each pixel in the image is typically determined by the imaging system or camera that captured the image. Eq (1) describes about the estimation of pixel size with the information of sensor size, number of pixels, distance of the sample object and the focal length of the camera.

\begin{equation}
\label{eq1}
PixelSize=\left(\frac{SensorSize}{No.ofPixels}\right)\times\left(\frac{DistToObject}{FocalLength}\right)
\end{equation}

\begin{equation}
\begin{aligned}
\label{eq2}
Area of Each Mask = Pixel Size \\ \times No.of Pixels in Each Mask
\end{aligned}
\end{equation}

The area of total identified debris in a frame involves calculating the area of each segmented mask of each class separately and then summing up the areas of all the masks to obtain the total area.

In cases where the object is complex, the Monte Carlo method is used to estimate the area. To estimate the area of an object using the Monte Carlo method, place 100 random points inside a rectangle of known area and count the number of points that lie within the object.
The area of the object is proportional to the number of points that lie inside it and is given by this formula: Eq (3).

\begin{equation}
\begin{aligned}
\label{eq3}
AreaofObject=AreaofRectangle \\ \times\frac{No.ofPointsInsideObject}{TotalNo.ofPointsinRectangle}
\end{aligned}
\end{equation}

Once the area is achieved, the volume of the marine debris is calculated by using standard formulas for regular shapes and for irregular shapes it is calculated using underwater Light Detection and Ranging (LiDAR) and photogrammetry. The computed mass is not the actually mass of the object, however they are helpful to identify an approximate value of debris deposited in the sea bed. Several other mass estimation techniques are also developed and applied for fish and coral reef biomass estimation. This aids cleaning robots to predetermine the capacity it can collect and carry back to the surface.  Mathematical models are also used to estimate the force required to move debris based on factors such as size, shape, water flow speed, viscosity, and density. Uncertainty may exist in these estimates due to factors such as water resistance and turbulence, which can be accounted for by including parameters such as the Reynolds number which describes the flow regime of the water, and the turbulence intensity in the water Eq (4).
\begin{equation}
\begin{aligned}
\label{eq4}
Re=\frac{\left(\rho\nu L\right)}{\mu}
\end{aligned}
\end{equation}

where Re is the Reynolds number, \(\rho\) is the density of the fluid, \(\nu\) is the velocity of the fluid, L is the characteristic length of the object or flow, and \(\mu\) is the dynamic viscosity of the fluid.
Encoders are used to measure a motor's rotational speed and calculate the torque being produced, which can be used to determine the force being applied to the debris. The torque is calculated using the formula Eq (5).

\begin{equation}
\begin{aligned}
\label{eq5}
Torque=k\times I\times\left(V-K'\times w\right)
\end{aligned}
\end{equation}

where k is a constant, I is the current flowing through the motor, V is the voltage applied to the motor, w is the motor's rotational speed measured in radians per second, and k' is a constant that depends on the motor's design.
Once the torque is calculated it is used to determine the force being applied to the debris using formula Eq (6)

\begin{equation}
\begin{aligned}
\label{eq5}
Force=\frac{Torque}{Radius}
\end{aligned}
\end{equation}

Acoustic Doppler Current Profilers ADCPs are used to detect and measure the velocity of water and the size and shape of underwater debris to estimate the mass of debris. The data is collected and preprocessed to remove noise and inconsistencies before feature extraction and model training.  For feature extraction Principal Component Analysis (PCA) is used which is a useful technique for dimensionality reduction, identifying patterns in large underwater datasets, for reducing noise and identifying the most important features in a dataset that helps to identify features such as the presence of marine life, effects of anthropogenic noise, and patterns in the distribution of objects or environmental changes. The technique involves transforming the data into a new coordinate system where the first principal component explains the maximum variation in the data, and each subsequent component explains the remaining variance in order of importance Testing and validating the instance segmentation model using a separate set of data to evaluate its performance. The performance metrics include mean Average Precision (mAP), mean Intersection over Union (mIoU), and F1-score.

Once the instance segmentation model is trained and validated, the data log is performed for every prediction by the AUV’s embedded system that interface with sensors and stores the data in memory. The XBee module equipped with a gateway device receives data from the XBee network and acts as a bridge between the local XBee network and the internet. It forwards data from the XBee network to the cloud platform and uses the cloud platform’s Application programming interface (API) to send the data to the cloud. The collected data is uploaded to the Google Cloud platform, which offers a range of tools for complex data analysis and processing, including BigQuery, Cloud Dataflow, and Cloud Dataproc. Uploading data to Google Cloud ensures that the data is backed up and can be recovered in case of data loss. Finally, the output is obtained in the form of a website for real-time monitoring of the computed data. Figure 1 illustrates the block diagram of the proposed methodology

\subsection{ESTIMATION OUTCOMES}
The volume and density are estimated for the debris found. The results are illustrated in Table 1.

\begin{table}
\begin{center}
\caption{Volume and density estimations of Marine debris}
\label{tab1}
\begin{tabular}{| p{1.2cm} | p{1.2cm} | p{1.2cm} | p{1.2cm} |}
\hline
\textbf{TYPES OF DEBRIS} & \textbf{DIMENSIONS (cm)} & \textbf{VOLUME (cm 3)} & \textbf{DENSITY (g/cm 3)}\\
\hline
Plastic beverage bottles & 6 x 6 x 23 & 829 & 1.4\\
\hline
Plastic bags & 10 x 7 x 4 & 280 & 1.2\\
\hline
Food containers & 16 x 12 x 7 & 1344 & 1.35\\
\hline
Glass beverage bottles & 7 x 7 x 26 & 1274 & 2.5\\
\hline
Fishing gear & 210 x 15 x 10 & 31500 & 7\\
\hline
Wood & 41 x 10 x 14 & 5740 & 6.5\\
\hline
Tyres & 19 x 10 x 40 & 7600 & 2.2\\
\hline
\end{tabular}
\end{center}
\end{table}
\begin{figure}[h]
\centering{\includegraphics[width=0.4\textwidth]{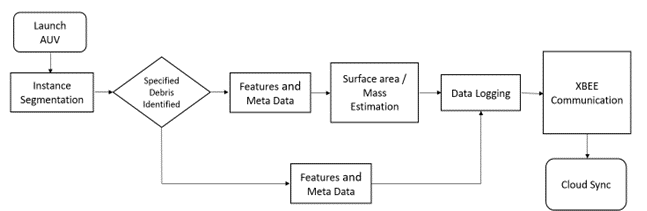}}
          \caption{Block Diagram of Proposed Methodology}
          \label{fig1}
          \end{figure}
\subsection{COMPUTER VISION}
Computer vision is dominating the current era and lots of research is being carried out by numerous researchers in this field. It is a field of artificial intelligence and computer science that focuses on enabling computers to interpret and understand visual information from the world. It involves developing algorithms and techniques to extract meaningful information from images and videos. Computer vision instructs machines to understand, grasp, and analyze a high-level understanding of visual contents. Its subfields include scene or object recognition, object detection, video tracking, object segmentation, pose and motion estimation, scene modeling, and image restoration. Leveraging the capabilities of computer vision, precise detection and estimation of the mass of marine debris can be achieved. By harnessing advanced image processing algorithms and pattern recognition techniques, computer vision enables the automated identification and categorization of various types of marine debris in aquatic environments. The application of computer vision in marine debris mass estimation not only expedites the data collection process but also minimizes human intervention, reducing potential biases and errors. This technology empowers researchers, environmentalists, and policymakers with real-time, data-driven insights to make informed decisions for marine conservation and protection. This paper focuses on the object detection, instance segmentation and its relevant subfields as the most important and popular tasks of computer vision.
\subsubsection{OBJECT DETECTION}
      Object detection is a significant field in the domain of computer vision. It plays a primary role in the use of Intelligent Debris Mass Estimation Model for Autonomous Underwater Vehicles (AUVs) enabling it to identify , accurately detect and localize debris object leading towards competent marine debris mass estimation model. In this paper amongst the object detection algorithms different versions of  YOLO algorithm is used for detecting and estimating the mass of marine debris as it posses the ability to manage images at a rapid rate of 45 Frames Per Second (FPS), achieving twice the average of mean Average Precision (mAP) and showing high-caliber detection accuracy comparing to other real-time systems making it a perfect choice for real-time processing. It works by the principle of dividing an image into a SxS grid of dimensions where each grid cell contains m bounding boxes. The network generates class probability and bounding box offset values within these bounding boxes. By choosing the bounding boxes with class probabilities above a specific threshold, the object is located within the image. It compares the detected debris instances with known debris categories and from expected debris types it identifies anomalies in the data.
\subsubsection{YOLOv3 }
      YOLO v3 introduced a new architecture called "Darknet-53", which consists  of 53 convolutional layers approving for finer feature extraction and representation of objects. It enables detection at discrete resolutions by making use of three different scales in the architecture. For smaller objects it further introduced the idea of "feature pyramid networks" (FPN) to collect information from different scales and to enhance detection performance. FPN let the model to detect dissimilar object sizes more efficiently by combining high-resolution features from early layers and low-resolution features from deeper layers. To improve accuracy it refined YOLO architecture and offered multi-scale detection. YOLO v3 achieved state-of-the-art accuracy with real-time detection. It developed a more resilient loss function called the focal loss to address the problem of class disproportion in object detection. YOLOv3 showed a remarkable enhancement in accuracy by illustrating a mAP of around 79-82\% on the Pascal VOC dataset compared to YOLOv2. It achieved an average precision AP of 36.2\% and AP50 of 60.6\% at 20 FPS making state-of-the-art during the time and 2× faster on MS COCO dataset [20].
\subsubsection{YOLOv4}
      YOLO v4 employed a modified CSPDarknet53 backbone architecture incorporating a Cross-Stage Partial Network (CSP) design to enhance feature extraction accuracy and speed. It brought forward the concept of the bag-of-freebies and the bag-of-specials to enhance model performance and Achieved state-of-the-art accuracy on the COCO benchmark when upholding real-time or near real-time performance. YOLO v4 introduced the Spatial Attention Module (SAM) to enhance feature representation and detection accuracy. It adopted diverse training strategies including DropBlock regularization, class-balanced loss, and focal loss, addressing generalization and challenging object detection. Advanced data augmentation techniques, like mosaic data augmentation and mixup augmentation, improved the model's adaptability to complex scenes and variations. Weighted-Residual-Connections (WRC) were introduced to enhance gradient flow and training convergence by employing weighted connections. YOLO v4 also proposed an ensemble approach, leveraging multiple models to enhance detection accuracy. It harnessed features like Weighted-Residual-Connections (WRC), Cross-Stage-Partial-connections (CSP), Cross mini-Batch Normalization (CmBN), Self-adversarial-training (SAT), Mish-activation and more to achieve state-of-the-art results, reaching 43.5\% AP (65.7\% AP50) on the MS COCO dataset at ~65 FPS on Tesla V100 in real-time [21].
\subsubsection{YOLOv5}
      YOLO v5 was developed independently inspired by the YOLO architecture. YOLOv5 is not endorsed by the original authors but it gained prominence for enhancing speed, simplicity, and accuracy in object detection. Its streamlined design incorporated a single network head that yielded quicker training and inference. YOLOv5 introduced an efficient model with anchor-free prediction and single-scale methodology . Instead of using predefined anchor boxes, YOLO v5 employed a CenterNet-style object detection approach directly. The integration of EfficientNet backbone further improved feature extraction and attained competitive accuracy .It considerably reduced the model size and inference duration. Evaluated on MS COCO dataset test-dev 2017, YOLOv5x achieved an AP of 50.7\% with an image size of 640 pixels. Using a batch size of 32, it can achieve a speed of 200 FPS on an NVIDIA V100. Using a larger input size of 1536 pixels, YOLOv5 achieves an AP of 55.8\% [22].
\subsubsection{YOLOv6}
      YOLOv6 stands out as the most accurate object detector, illustrated by YOLOv6 Nano achieving a 35.6\% mAP on COCO dataset and sustaining over 1200 FPS on NVIDIA Tesla T4 Graphics processing unit (GPU) with a batch size of 32. The achievement is attributed to novel approaches like reparameterized backbones, model quantization, and diverse augmentations. Unlike its predecessors, YOLOv6 employs an anchor-free method for object detection, enhancing generalization and reducing post-processing time. The model architecture features a revamped reparameterized backbone and neck, utilizing Varifocal loss (VFL) for classification and Distribution Focal loss (DFL) for detection. YOLOv6's strategies, like prolonged training, quantization, and knowledge distillation, render it optimal for real-time industrial use, boasting 51\% faster speed due to significantly fewer priors. The EfficientRep backbone, comprising RepBlock, RepConv, and CSPStackRep blocks, underpins YOLOv6 [23].
\subsubsection{YOLOv7}
      YOLOv7 is the fastest and most accurate real-time object detection model for computer vision tasks. The model is important for distributed real-world computer vision applications. The integration of YOLOv7 with BlendMask is used to perform instance segmentation. Therefore, the YOLOv7 object detection model was fine-tuned on the MS COCO instance segmentation dataset and trained for 30 epochs. YOLOv7 provides a greatly improved real-time object detection accuracy without increasing the inference costs. YOLOv7 surpasses all previous object detectors in terms of both speed and accuracy, ranging from 5 FPS to as much as 160 FPS. The YOLOv7 algorithm achieves the highest accuracy among all other real-time object detection models while achieving 30 FPS or higher using a GPU V100.   Comparison with other real-time object detectors YOLOv7 achieves state-of-the-art (SOTA) performance. Source Compared to the best performing Cascade-Mask R-CNN models, YOLOv7 achieves 2\% higher accuracy at a dramatically increased inference speed (509\% faster). This is impressive because such Region-based Convolutional Neural Network(R-CNN) versions use multi-step architectures that previously achieved significantly higher detection accuracies than single-stage detector architectures. YOLOv7 outperforms YOLOR, YOLOX, Scaled-YOLOv4, YOLOv5, End-to-End Object Detection with Transformers (DETR), Vision Transformers (ViT), Adapter-B, and many more object detection algorithms in speed and accuracy [24].
\subsubsection{YOLOv8}
      YOLOv8 is a versatile deep learning model introduced by Ultralytics. YOLOv8 offers capabilities in object detection, instance segmentation and image classification. Its training speed exceeds the  conventional two-stage object detection models. A distinctive feature of YOLOv8 is its anchor-free design. It reduces the volume of box predictions and quickens Non-maximum Suppression (NMS) processing. YOLOv8 employs mosaic augmentation during training. Due to future drawbacks it is  deactivated in the final ten epochs. YOLOv8 model can be operated via command line interface (CLI) or can be installed as a PIP package. It  includes various integrations for tasks such as labeling, training, and deployment. YOLOv8 offers five scaled variants: YOLOv8n (nano), YOLOv8s (small), YOLOv8m (medium), YOLOv8l (large), and YOLOv8x (extra large). YOLOv8x achieved an AP of 53.9\% at an image size of 640 pixels comparing to YOLOv5's 50.7\% on the same input size. In the experiments conducted on the MS COCO dataset test-dev 2017 it achieved a speed of 280 FPS on an NVIDIA A100 and TensorRT [25].
\subsection{DRAWBACKS }
Accurate measurement of object size is necessary for estimating the mass in realtion with Object detection. The grid-based methodology of  YOLO resulted in uncertain localization and object boundaries specifically for complex shaped debris. Debris overlaped within a single grid cell resulted in incorrect detection of merged debris as it has the possibility to accumulate and cluster in the marine realm. It detects using a single bounding box that may envelops both objects making it difficult to differentiate and classify individual objects. Due to its grid structure it struggled to detect tiny debris and not adequately capture fine details and classify them. The aspect ratios of the fixed bounding box did not adequately handle the objects with extreme aspect ratios. So in some cases YOLO's real-time focus prioritized speed over accuracy.
\subsection{INSTANCE SEGMENTATION}
Instance segmentation is pivotal in the Intelligent Debris Mass Estimation Model for Autonomous Underwater Vehicles (AUVs) for accurately detecting, segmenting, and estimating debris in underwater settings. Unlike conventional object detection, this advanced computer vision technique not only identifies objects but also precisely outlines individual object boundaries. Instance segmentation algorithms such as YOLO v5, YOLOv7, and YOLOv8  effectively locate debris instances in captured images. These algorithms are trained on extensive annotated underwater datasets and use deep learning to recognize diverse debris types. In contrast to traditional approach, instance segmentation provides pixel-level classification which is crucial for precise mass estimation by delineating object shapes, including irregularities and overlaps. YOLO instance segmentation ensures precise object boundaries and spatial arrangements. Overlapping objects are handled adeptly by YOLO, distinguishing individual instances in the same area. Segmentation masks are created by applying binary masks and thresholds. They are refined through post-processing for improved accuracy [26].
\subsubsection{YOLOv5}
      The YOLOv5 instance segmentation models are renowned for remarkable speed and precision in real-time instance segmentation tasks. It comprises of an object detection head and the ProtoNet to generate prototype masks for segmentation, similar to an FCN with SiLU activations. Detection layers operate at three different scales, each yielding three anchor boxes and ProtoNet outputs prototype masks. The  final convolutional detection heads have 351 channels differing from the standard 255. To secure confinement masks are clipped to bounding. YOLOv5 Extra Large (yolov5x-seg) achieves 41.4 mask mAP on A100 GPU with TensorRT running at 833 FPS with 1.2ms latency being the top-performing model. All models trained for 300 epochs on COCO dataset using NVIDIA A100 GPU. YOLOv5 exceeds ResNet101-backed models by perfoming faster.
\subsubsection{YOLOv7}
      The YOLOv7 object detection model underwent fine-tuning using the MS COCO instance segmentation dataset and their training extended to 30 epochs. The combination of YOLOv7 and BlendMask enhances the  instance segmentation capabilities of YOLOv7. This training process yielded cutting-edge real-time instance segmentation results. The incorporation of YOLOv7 into YOLO-Pose opens up opportunities for keypoint detection in Pose Estimation. The further refinement of YOLOv7-W6 model focused in detecting people in the MS COCO keypoint detection dataset and obtained state-of-the-art real-time pose estimation performance.
\subsubsection{YOLOv8}
      The YOLOv8-Seg model is an evolution of the YOLOv8 object detection model that performs instance segmentation of the provided image.  The CSPDarknet53 feature extractor is the foundation of the YOLOv8-Seg model at its core. In place of the traditional YOLO neck architecture this model adopted a novel C2f module. Two segmentation heads are responsible for predicting the instance segmentation masks for the given image. Similar to YOLOv8 this model  contains five detection modules and a prediction layer. The YOLOv8-Seg model demonstrated state-of-the-art results while maintaining high speed and efficiency.
\section{EXPERIMENTS AND RESULTS}
To measure the yolo model’s performance, several sets of experiments were carried out on a trash-can dataset of images containing variety of objects. To ensure optimum efficiency for object detection and instance segmentation tasks the models were configured with clear-cut backbone architecture and hyperparameters. Certain hyperparameters like learning rate, batch size, and regularization parameters were carefully chosen to impact the model’s learning. The dataset was cautiously annotated that displayed classes with precise bounding box coordinates and their respective class labels. On the basis of objects shapes and sizes in the dataset the anchor box sizes and aspect ratios were selected. Data Augmentation Techniques such as cropping, flipping, rotation, and others were applied to images in each batch before inputting it to the model to enhance the model's generalization by introducing a broad range of variations. The dataset was iteratively processed multiple times using optimization algorithms, including stochastic gradient descent, to fine-tune the model's weights. To monitor the model's performance throughout training, the dataset’s subset was set up for validation purpose. If the performance on the validation set began to downturn during the process, the training was halted early to avoid overfitting. After training  the models' performance are evaluated using metrics - Precision (P), Recall (R), mean Average Precision mAP@0.5, mAP@0.5 :0.95 and F1 score. These metrics measure the model's efficiency to accurately identify and segment objects within the images refining their real-world relevance.
\subsection{DATASET AND ANNOTATIONS}
The [27] TrashCan dataset, introduced by Hong et al.,was utilized for the study and annotated using Roboflow, as it is a robust platform for annotating datasets for various computer vision tasks. The dataset images were uploaded and annotation process started by creating masks, polygons and bounding boxes to label the objects. To verify whether the images are properly prepared, image preprocessing techniques like resizing and augmentation are applied. For annotating bounding boxes around the objects are drawn in object detection. For instance segmentation, annotation tools like masks or polygons is used for precisely tracing the edges of each individual object in the image. The mask for each instance is a binary image, where  pixels within the mask correspond to the object, while pixels outside the mask depict the background. After the annotation process class labels are assigned to each annotated objects. This step determines the class of each object, helping the model in learning to distinguish between different object classes. The annotation accuracy is reviewed and verified by validation process. Once the process is complete, the annotated dataset is exported in compatible formats such as YOLO Darknet, YOLO V3 Keras, YOLO V4 Pytorch, YOLO V5 Pytorch, , YOLO V7 Pytorch and YOLOV8. The exported datatset is integrated into the YOLO Algorithms and the model attained 97.2\% mAP.
\subsection{OBJECT DETECTION}

\subsubsection{YOLOv3}
      YOLOv3 demonstrated precision(P) and recall(R) scores of 0.9028 and 0.954 respectively. This dual success emphasized the model's ability to minimize false positives while capturing a substantial number of true positives. YOLOv3 showcased an impressive mAP score, attesting to its efficiency in detecting trash cans across diverse scenarios. At an IoU threshold of 0.5, the model achieved an mAP@0.5 of 0.9632. Over a range of IoU thresholds from 0.5 to 0.95, it attained 0.7822, highlighting its adaptability to various object overlaps. The F1 score of 0.9276 underscored YOLOv3's equilibrium between precision and recall, encapsulating its comprehensive detection prowess. The results of object detection using YOLOv3 are displayed in Figure 2.

\begin{table*}[!t]
\begin{center}
\caption{YOLOv3 Object detection experiment outcome}
\label{tab2}
\begin{tabular}{| p{2.0cm} | p{2.0cm} | p{2.0cm} | p{2.0cm} | p{2.0cm} | p{2.0cm} | p{2.0cm} |}
\hline
\textbf{Class} & \textbf{Images} & \textbf{Labels} & \textbf{P} & \textbf{R} & \textbf{mAP@.5} & \textbf{mAP@.5:95
}\\
\hline
All & 117 & 152 & 0.903 & 0.954 & 0.963 & 0.783
\\
\hline
Crab & 117 & 16 & 0.843 & 1 & 0.988 & 0.832
\\
\hline
Fish & 117 & 37 & 0.983 & 1 & 0.995 & 0.814
\\
\hline
Machines & 117 & 58 & 0.914 & 0.914 & 0.944 & 0.737
\\
\hline
Trash & 117 & 41 & 0.871 & 0.902 & 0.926 & 0.748
\\
\hline
\end{tabular}
\end{center}
\end{table*}
\begin{figure}[h]
\centering{\includegraphics[width=0.4\textwidth]{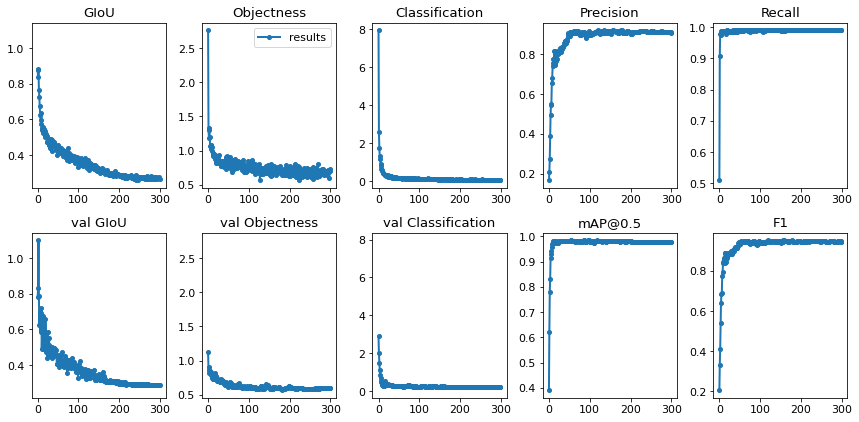}}
          \caption{YOLOv3 Object detection graph outcome}
          \label{fig2}
          \end{figure}
\subsubsection{YOLOv5     }
      YOLOv5 exhibited precision(P) and recall(R) results of 0.965 and 0.9602, respectively. This dual accomplishment highlighted the model's capacity to minimize false positives while effectively capturing a substantial count of true positives. YOLOv5 demonstrated an impressive mean Average Precision (mAP) affirming its effectiveness in identifying trash cans across a wide array of scenarios. At an Intersection over Union (IoU) threshold of 0.5, the model attained an mAP@0.5 of 0.9668. Across a range of IoU thresholds spanning from 0.5 to 0.95, it achieved a score of 0.804, showcasing its flexibility in accommodating various levels of object overlap. The F1 score of 0.9624 underscored YOLOv5's equilibrium between precision and recall, encapsulating its comprehensive detection capabilities.

\begin{table*}[!t]
\begin{center}
\caption{YOLOv5 Object detection experiment outcome}
\label{tab3}
\begin{tabular}{| p{2.0cm} | p{2.0cm} | p{2.0cm} | p{2.0cm} | p{2.0cm} | p{2.0cm} | p{2.0cm} |}
\hline
\textbf{Class} & \textbf{Images} & \textbf{Labels} & \textbf{P                   } & \textbf{R} & \textbf{mAP@.5} & \textbf{mAP@.5:95
}\\
\hline
All & 117 & 152 & 0.972 & 0.959 & 0.967 & 0.804
\\
\hline
Crab & 117 & 16 & 0.94 & 1 & 0.991 & 0.841
\\
\hline
Fish & 117 & 37 & 0.993 & 1 & 0.991 & 0.837
\\
\hline
Machines & 117 & 58 & 0.952 & 0.944 & 0.962 & 0.798
\\
\hline
Trash & 117 & 41 & 0.968 & 0.898 & 0.923 & 0.740
\\
\hline
\end{tabular}
\end{center}
\end{table*}
\subsubsection{YOLOv7}
      YOLOv7 displayed precision and recall scores of 0.9762 and 0.9626 correspondingly. This twofold achievement highlighted the model's aptitude for minimizing false positives while successfully capturing a significant number of true positives. YOLOv7 presented an exceptional mean Average Precision (mAP) score, affirming its efficacy in recognizing trash cans across diverse scenarios. With an Intersection over Union (IoU) threshold of 0.5, the model accomplished an mAP@0.5 of 0.9708. Spanning a range of IoU thresholds from 0.5 to 0.95, it achieved a value of 0.808, underscoring its versatility in accommodating varying degrees of object overlap. The F1 score of 0.9692 highlighted YOLOv7's balance between precision and recall, encapsulating its all-encompassing detection prowess. The outcome of object detection with YOLOv7 is presented in Figure 3.

\begin{table*}[!t]
\begin{center}
\caption{YOLOv7 Object detection experiment outcome}
\label{tab4}
\begin{tabular}{| p{2.0cm} | p{2.0cm} | p{2.0cm} | p{2.0cm} | p{2.0cm} | p{2.0cm} | p{2.0cm} |}
\hline
\textbf{Class} & \textbf{Images} & \textbf{Labels} & \textbf{P                   } & \textbf{R} & \textbf{mAP@0.5} & \textbf{  mAP@0.5 :95
}\\
\hline
All & 117 & 152 & 0.976 & 0.963 & 0.971 & 0.808
\\
\hline
Crab & 117 & 16 & 0.98 & 1 & 0.995 & 0.845
\\
\hline
Fish & 117 & 37 & 0.997 & 1 & 0.995 & 0.841
\\
\hline
Machines & 117 & 58 & 0.956 & 0.948 & 0.966 & 0.802
\\
\hline
Trash & 117 & 41 & 0.972 & 0.902 & 0.927 & 0.744
\\
\hline
\end{tabular}
\end{center}
\end{table*}
\begin{figure}[h]
\centering{\includegraphics[width=0.4\textwidth]{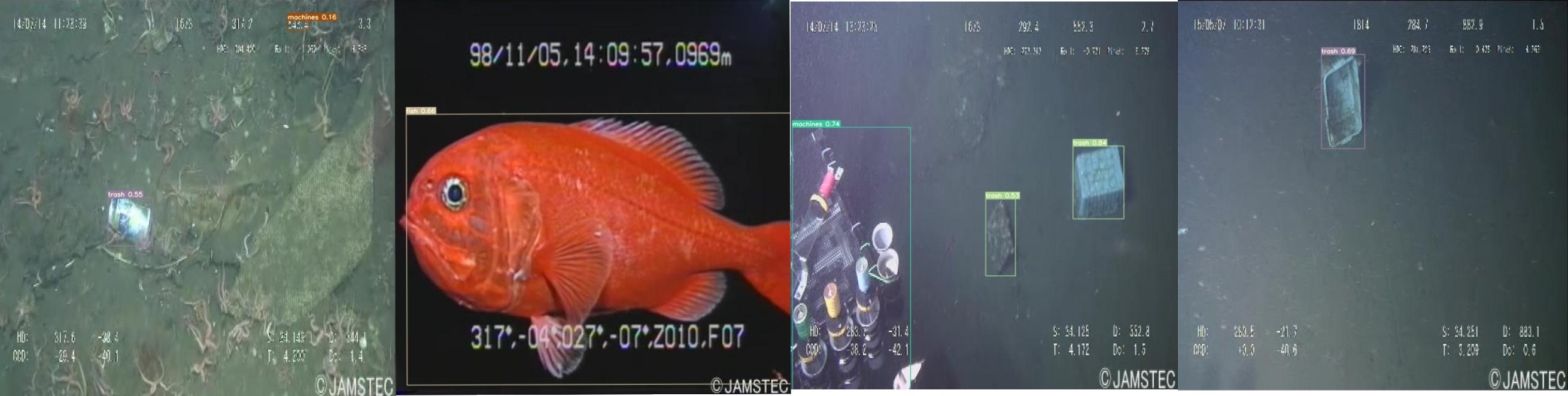}}
          \caption{YOLOv7 Object detected outcome}
          \label{fig3}
          \end{figure}
\subsubsection{YOLOv8}
      The YOLOv8 model showcased impressive precision and recall metrics, with a precision (P) score of 0.913 and a recall (R) score of 0.9548. These values underscore the model's precision and its capacity to effectively reduce both false positives and false negatives. YOLOv8's performance was especially noteworthy in terms of the mean average precision, where it achieved an outstanding mAP score. To be specific, at an intersection over union (IoU) threshold of 0.5, the mAP@0.5 score stood at 0.967. Furthermore, across a range of IoU thresholds from 0.5 to 0.95, the mAP@0.5:0.95 score reached an impressive 0.832, highlighting its consistent and dependable performance across varying levels of object overlap. The F1 score, a well-balanced metric considering both precision and recall, reached 0.9333. This accomplishment validates the model's ability to strike a harmonious balance between these two critical aspects, demonstrating its comprehensive and adaptable performance. The graphical representation of the object detection results obtained using YOLOv8 is showcased in Figure 4. The results of object detection using YOLOv8 are showcased in Figure 5.

\begin{table*}[!t]
\begin{center}
\caption{YOLOv8 Object detection experiment outcome}
\label{tab5}
\begin{tabular}{| p{2.0cm} | p{2.0cm} | p{2.0cm} | p{2.0cm} | p{2.0cm} | p{2.0cm} | p{2.0cm} |}
\hline
\textbf{Class} & \textbf{Images} & \textbf{Labels} & \textbf{P                   } & \textbf{R} & \textbf{mAP50       } & \textbf{  mAP50-95
}\\
\hline
All & 117 & 152 & 0.913 & 0.955 & 0.967 & 0.832
\\
\hline
Crab & 117 & 16 & 0.874 & 1 & 0.978 & 0.82
\\
\hline
Fish & 117 & 37 & 0.992 & 1 & 0.995 & 0.92
\\
\hline
Machines & 117 & 58 & 0.837 & 0.914 & 0.947 & 0.713
\\
\hline
Trash & 117 & 41 & 0.949 & 0.905 & 0.948 & 0.876
\\
\hline
\end{tabular}
\end{center}
\end{table*}
\begin{figure}[h]
\centering{\includegraphics[width=0.4\textwidth]{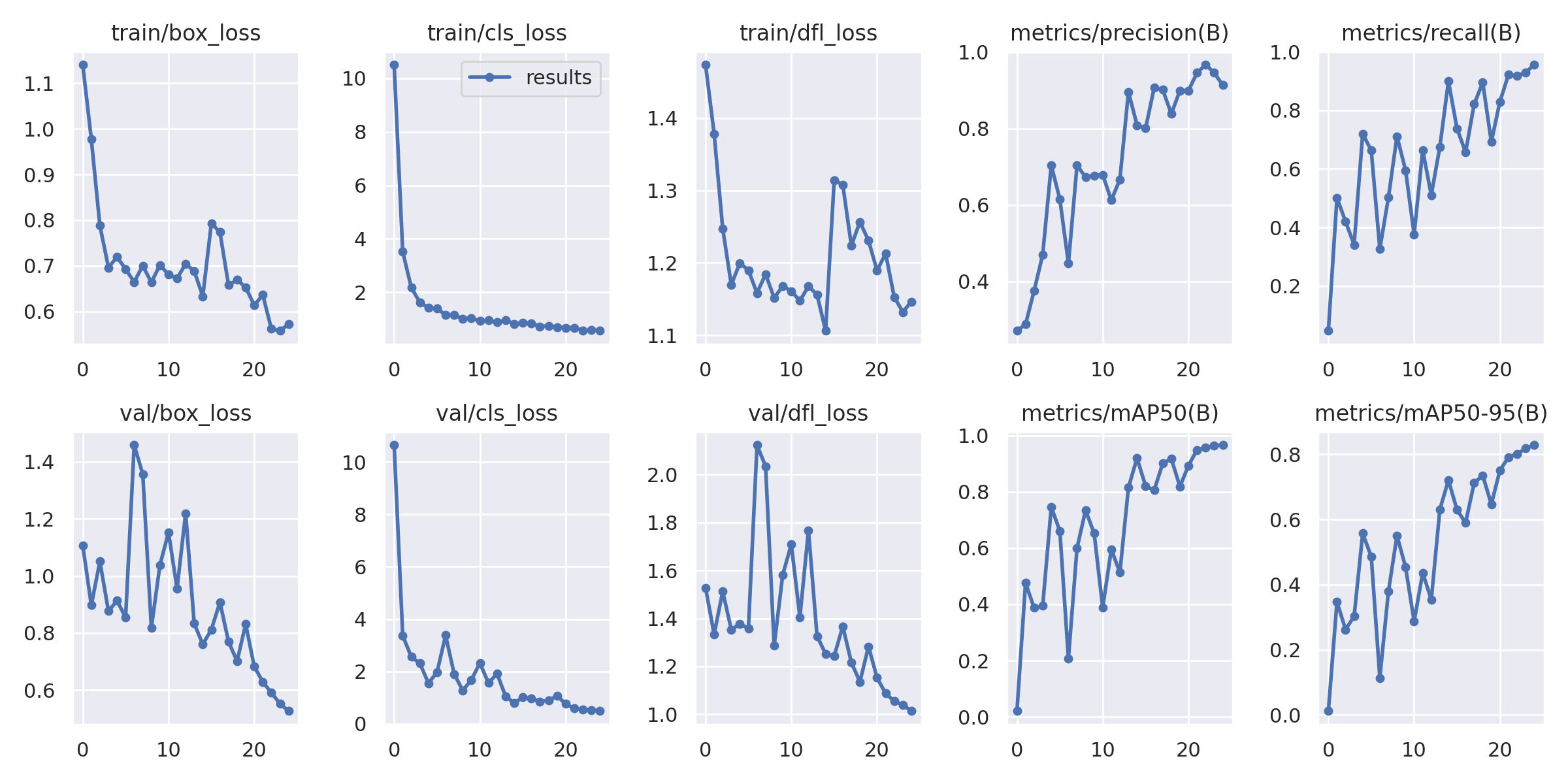}}
          \caption{YOLOv8 Object detection graph outcome}
          \label{fig4}
          \end{figure}
\begin{figure}[h]
\centering{\includegraphics[width=0.4\textwidth]{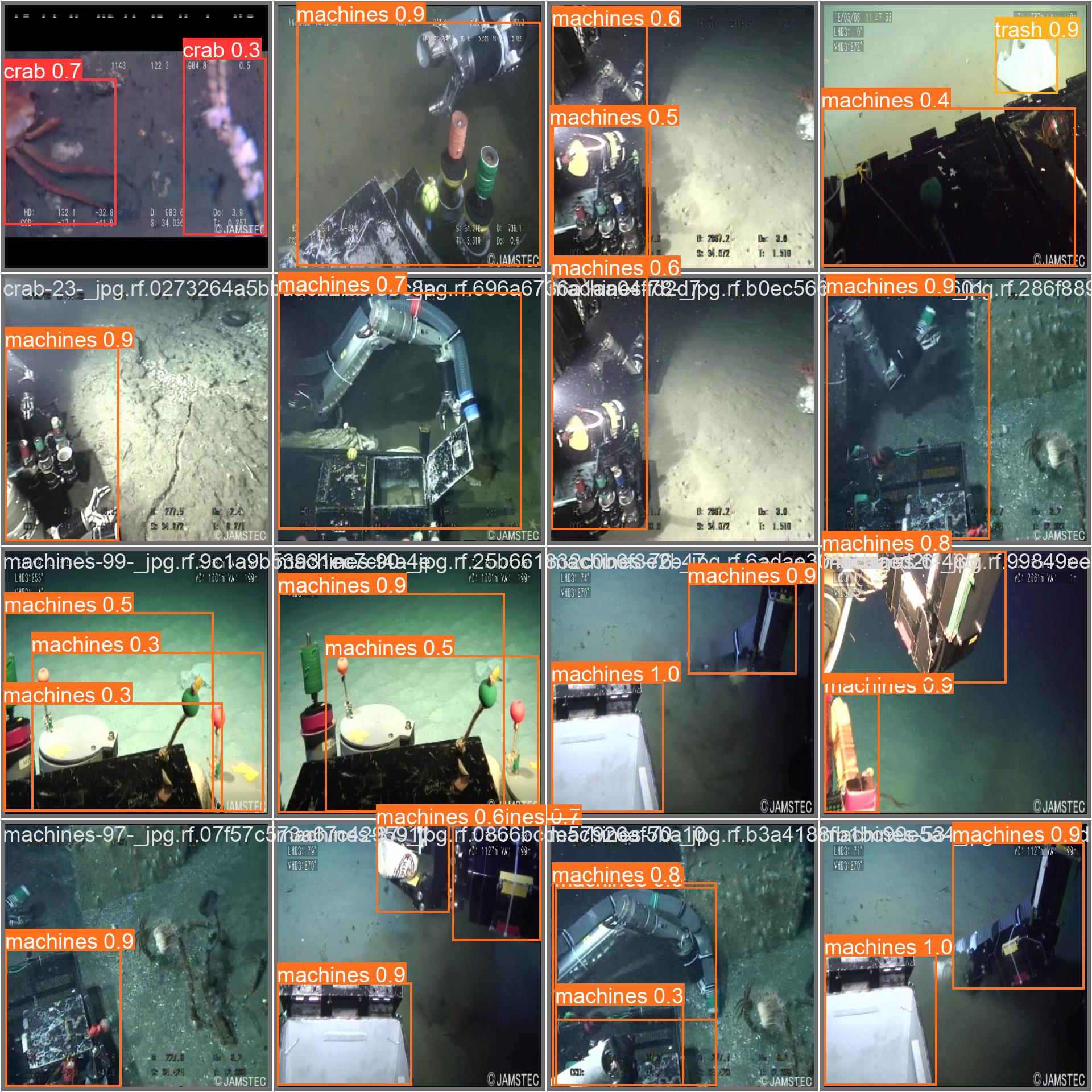}}
          \caption{YOLOv8 Object detected outcome}
          \label{fig5}
          \end{figure}
\subsection{INSTANCE SEGMENTATION }

\subsubsection{ YOLOv5}
      The YOLOv5 instance segmentation model exhibited impressive precision and recall outcomes. It attained a precision (P) score of 0.9762 and a recall (R) score of 0.9626, emphasizing its accuracy in minimizing false positives and false negatives. YOLOv5's performance excelled in mean average precision evaluation, achieving a notable mAP score. Notably, at an intersection over union (IoU) threshold of 0.5, the mAP@0.5 score reached 0.9722. Moreover, across IoU thresholds spanning from 0.5 to 0.95, the mAP@0.5:0.95 demonstrated consistency by achieving 0.7574 underscoring its reliability across diverse object overlap scenarios. The F1 score, a holistic metric that combines precision and recall reached 0.9692. This success further underscores the model's capacity to strike a harmonious balance between precision and recall, accentuating its adaptability and all-encompassing performance. Figure 6 illustrates the findings of Instance segmentation achieved through YOLOv5.
\begin{figure}[h]
\centering{\includegraphics[width=0.4\textwidth]{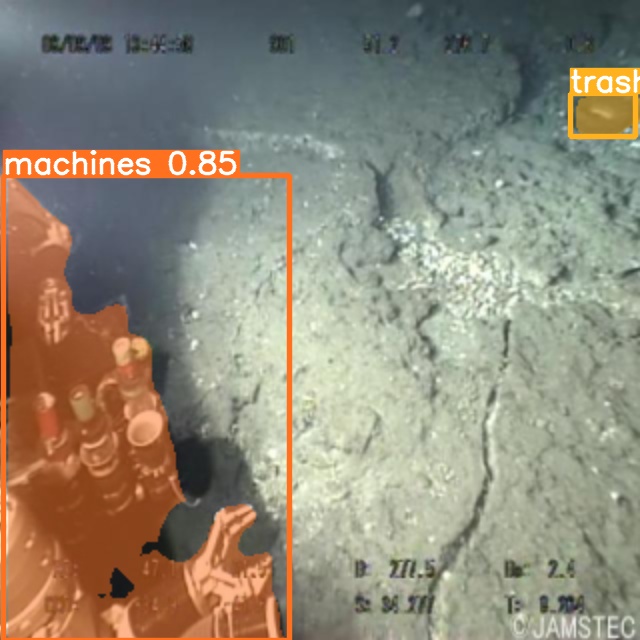}}
          \caption{YOLOv5 Instance segmented outcome}
          \label{fig6}
          \end{figure}

\begin{table*}[!t]
\begin{center}
\caption{YOLOv5 instance segmentation experiment outcome}
\label{tab6}
\begin{tabular}{| p{1.3cm} | p{1.3cm} | p{1.3cm} | p{1.3cm} | p{1.3cm} | p{1.3cm} | p{1.3cm} | p{1.3cm} | p{1.3cm} | p{1.3cm} | p{1.3cm} |}
\hline
\textbf{Class} & \textbf{Images} & \textbf{Labels} & \multicolumn{4}{|c}{\textbf{Box}} & \multicolumn{4}{c|}{\textbf{Mask}}\\
\hline
\multicolumn{3}{|c|}{Parameters} & P & R & mAP @ 50 & mAP@50:95 & P & R & mAP@50 & mAP@50:95
\\
\hline
All & 117 & 152 & 0.971 & 0.963 & 0.971 & 0.808 & 0.976 & 0.963 & 0.972 & 0.757
\\
\hline
Crab & 117 & 16 & 0.995 & 1 & 0.995 & 0.845 & 0.98 & 1 & 0.995 & 0.691
\\
\hline
Fish & 117 & 37 & 0.995 & 1 & 0.995 & 0.841 & 0.997 & 1 & 0.995 & 0.835
\\
\hline
Machines & 117 & 58 & 0.966 & 0.948 & 0.966 & 0.802 & 0.956 & 0.948 & 0.966 & 0.726
\\
\hline
Trash & 117 & 41 & 0.927 & 0.902 & 0.927 & 0.744 & 0.972 & 0.902 & 0.933 & 0.778
\\
\hline
\end{tabular}
\end{center}
\end{table*}
\subsubsection{YOLOv7}
      The YOLOv7 instance segmentation model yielded impressive outcomes concerning precision and recall. It secured a precision (P) rating of 0.632 and a recall (R) score of 0.726, illustrating its precision in minimizing both false positives and false negatives. YOLOv7's prowess was most evident in its mean average precision (mAP) evaluation, where it achieved a remarkable mAP score. Specifically, at an intersection over union (IoU) threshold of 0.5, the mAP@0.5 score reached 0.7218. Furthermore, across IoU thresholds spanning from 0.5 to 0.95, the mAP@0.5:0.95 consistently demonstrated reliability by attaining 0.517, spotlighting its steadfastness across various scenarios of object overlap. The F1 score, an all-encompassing metric that amalgamates precision and recall, achieved 0.6756. This achievement further underscores the model's capacity to maintain a harmonious equilibrium between precision and recall, demonstrating its versatility and comprehensive performance. The depiction of YOLOv7 object detection results can be observed in Figure 7.

\begin{table*}[!t]
\begin{center}
\caption{YOLOv7 instance segmentation experiment outcome}
\label{tab7}
\begin{tabular}{| p{1.3cm} | p{1.3cm} | p{1.3cm} | p{1.3cm} | p{1.3cm} | p{1.3cm} | p{1.3cm} | p{1.3cm} | p{1.3cm} | p{1.3cm} | p{1.3cm} |}
\hline
\textbf{Class} & \textbf{Images} & \textbf{Labels} & \multicolumn{4}{|c}{\textbf{Box}} & \multicolumn{4}{c|}{\textbf{Mask}}\\
\hline
\multicolumn{3}{|c|}{Parameters} & P & R & mAP @ 50 & mAP@50:95 & P & R & mAP@50 & mAP@50:95
\\
\hline
All & 117 & 152 & 0.635 & 0.73 & 0.725 & 0.469 & 0.632 & 0.726 & 0.722 & 0.517
\\
\hline
Crab & 117 & 16 & 0.812 & 0.375 & 0.612 & 0.408 & 0.812 & 0.375 & 0.612 & 0.411
\\
\hline
Fish & 117 & 37 & 0.417 & 0.838 & 0.664 & 0.386 & 0.417 & 0.838 & 0.659 & 0.454
\\
\hline
Machines & 117 & 58 & 0.698 & 0.879 & 0.805 & 0.562 & 0.684 & 0.862 & 0.798 & 0.504
\\
\hline
Trash & 117 & 41 & 0.615 & 0.829 & 0.818 & 0.521 & 0.615 & 0.829 & 0.818 & 0.699
\\
\hline
\end{tabular}
\end{center}
\end{table*}
\begin{figure}[h]
\centering{\includegraphics[width=0.4\textwidth]{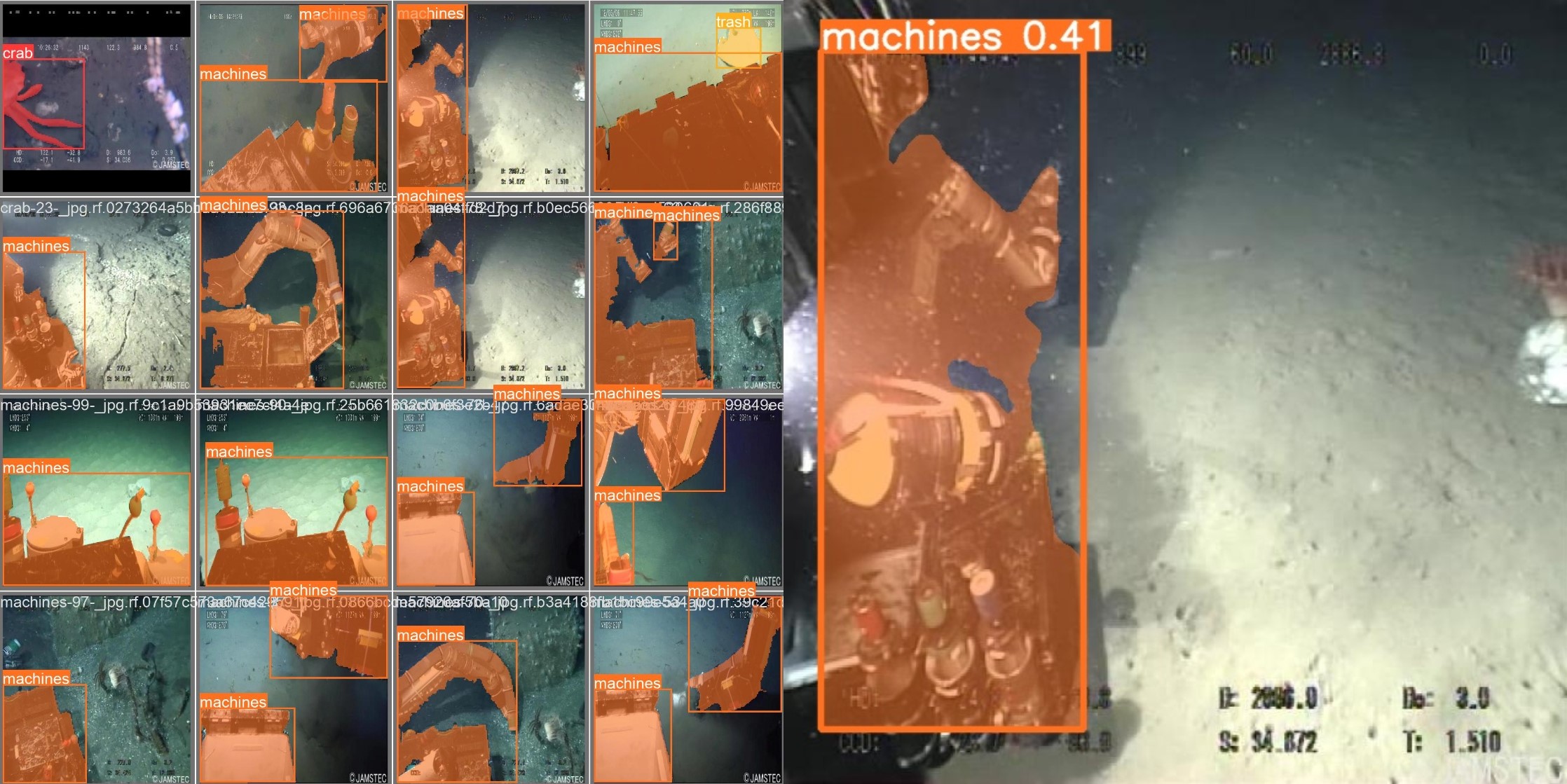}}
          \caption{YOLOv7 Instance segmented outcome}
          \label{fig7}
          \end{figure}
\subsubsection{ YOLOv8 }
      The YOLOv8 instance segmentation model delivered impressive precision and recall results. It achieved a precision (P) score of 0.908 and a recall (R) score of 0.915, emphasizing its precision in effectively minimizing both false positives and false negatives. YOLOv8's excellence was particularly evident in the mean average precision (mAP) assessment, where it secured a noteworthy mAP score. In detail, at an intersection over union (IoU) threshold of 0.5, the model attained an mAP@0.5 score of 0.955. Furthermore, across a range of IoU thresholds spanning from 0.5 to 0.95, the mAP@0.5:0.95 consistently demonstrated its reliability by achieving 0.758, highlighting its consistent performance across diverse scenarios of object overlap. The F1 score, an all-encompassing metric that combines precision and recall, reached 0.9114. This accomplishment further underscores the model's ability to maintain a harmonious equilibrium between precision and recall, accentuating its adaptability and comprehensive performance. The representation of the results from YOLOv8 object detection is depicted in Figure 8.

\begin{table*}[!t]
\begin{center}
\caption{YOLOv8 Instance segmentation experiment outcome}
\label{tab8}
\begin{tabular}{| p{1.3cm} | p{1.3cm} | p{1.3cm} | p{1.3cm} | p{1.3cm} | p{1.3cm} | p{1.3cm} | p{1.2cm} | p{1.2cm} | p{1.2cm} | p{1.2cm} |}
\hline
\textbf{Class} & \textbf{Images} & \textbf{Labels} & \multicolumn{4}{|c}{\textbf{Box}} & \multicolumn{4}{c|}{\textbf{Mask}}\\
\hline
\multicolumn{3}{|c|}{Parameters} & P & R & mAP @ 50 & mAP@50:95 & P & R & mAP@50 & mAP@50:95
\\
\hline
All & 117 & 152 & 0.906 & 0.901 & 0.95 & 0.828 & 0.908 & 0.915 & 0.955 & 0.758
\\
\hline
Crab & 117 & 16 & 0.849 & 1 & 0.977 & 0.843 & 0.846 & 1 & 0.977 & 0.645
\\
\hline
Fish & 117 & 37 & 0.984 & 0.973 & 0.994 & 0.92 & 0.983 & 0.973 & 0.994 & 0.875
\\
\hline
Machines & 117 & 58 & 0.892 & 0.828 & 0.931 & 0.763 & 0.911 & 0.882 & 0.95 & 0.71
\\
\hline
Trash & 117 & 41 & 0.898 & 0.805 & 0.899 & 0.786 & 0.892 & 0.805 & 0.899 & 0.804
\\
\hline
\end{tabular}
\end{center}
\end{table*}
\begin{figure}[h]
\centering{\includegraphics[width=0.4\textwidth]{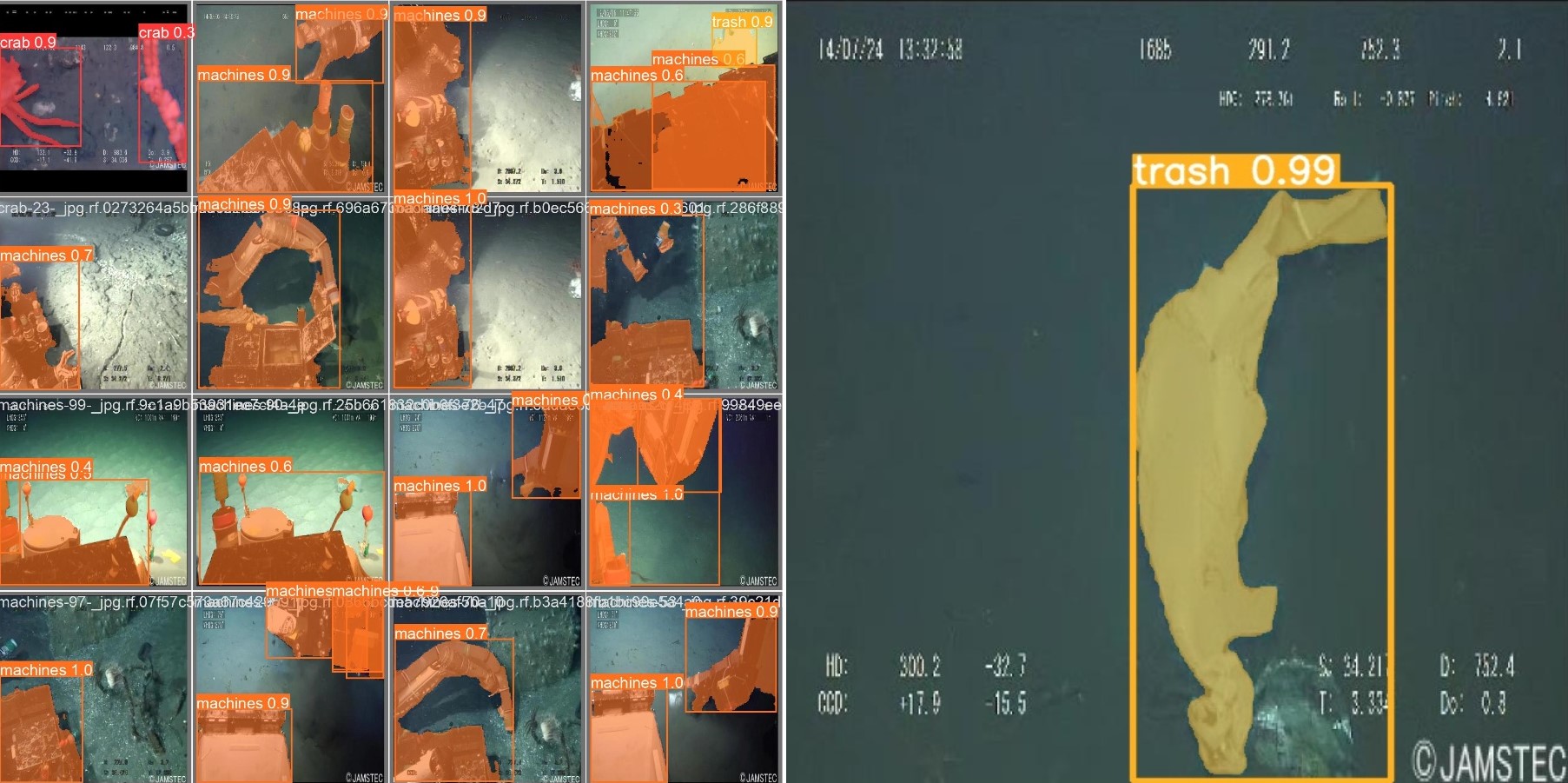}}
          \caption{YOLOv8 Instance segmented outcome}
          \label{fig8}
          \end{figure}
\section{CONCLUSION}
In this paper different versions of YOLO object detection and instance segmentation models performance and their respective Precision (P), Recall (R), mean Average Precision (mAP)@0.5, mAP@0.5 :0.95 and F1 score are obtained . The comparative study of these models accuracy are done in which the instance segmentation models provided better accuracy in detecting marine debris making it a robust model to precisely estimate the mass of the marine debris.  By using the Monte Carlo method, underwater LiDAR , photogrammetry, standard formulas for volume and density the mass of the marine debris is obtained. The Reynolds number, Archimedes principle , Sutherland’s equation are calculated for obtaining the torque value to pull the load and it is given to the motor. These collected data are stored in the Google Cloud and the data obtained are presented in the website format. Figure 9 illustrates the output of the proposed website.
\begin{figure}[h]
\centering{\includegraphics[width=0.4\textwidth]{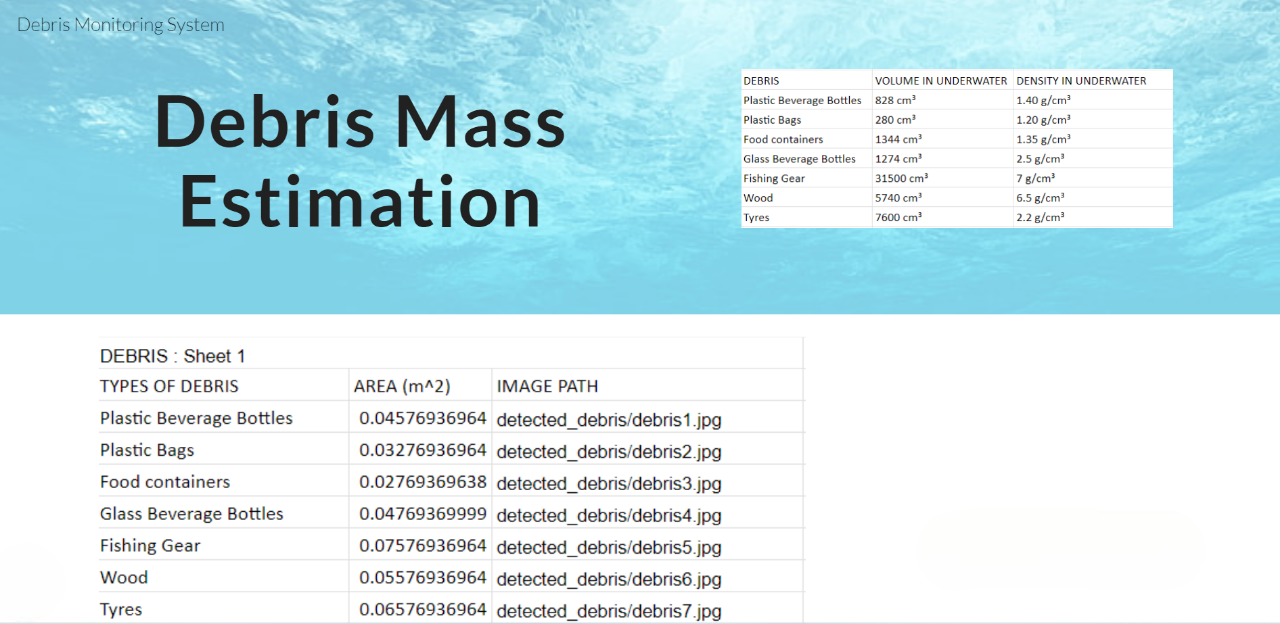}}
          \caption{Outcome of the Proposed Website}
          \label{fig9}
          \end{figure}

\FloatBarrier

\end{document}